\documentclass[10pt,twocolumn,letterpaper]{article}

\usepackage[pagenumbers]{cvpr} 

\usepackage{newfloat}
\usepackage{listings}
\usepackage{amsmath}
\usepackage{graphicx}
\usepackage{colortbl}
\usepackage{xspace}
\usepackage{bbding}
\usepackage{amssymb}
\usepackage{color}
\usepackage{arydshln}

\definecolor{cvprblue}{rgb}{0.21,0.49,0.74}
\usepackage[pagebackref,breaklinks,colorlinks,allcolors=cvprblue]{hyperref}

\def\paperID{10943} 
\def\confName{CVPR}
\def\confYear{2025}

\title{One Model for ALL: Low-Level Task Interaction Is a \\ Key to Task-Agnostic Image Fusion}

\author{Chunyang Cheng\textsuperscript{1}, Tianyang Xu\textsuperscript{1},  Zhenhua Feng\textsuperscript{1}, Xiaojun Wu\textsuperscript{1}\thanks{Corresponding Author}, ZhangyongTang\textsuperscript{1}, Hui Li\textsuperscript{1},\\Zeyang Zhang\textsuperscript{1},
Sara Atito\textsuperscript{2}, Muhammad Awais\textsuperscript{2}, Josef Kittler\textsuperscript{2}\\
\small\textsuperscript{1}School of Artificial Intelligence and Computer Science, Jiangnan University, Wuxi, China\\
\small\textsuperscript{2}Centre for Vision, Speech and Signal Processing (CVSSP), University of Surrey, Guildford, UK\\
{\tt\small \{chunyang\_cheng, zhangyong\_tang, zeyang\_zhang\}@stu.jiangnan.edu.cn} \\
{\tt\small \{tianyang\_xu, fengzhenhua, wu\_xiaojun, lihui.cv\}@jiangnan.edu.cn}
\\
{\tt\small \{sara.atito, muhammad.awais, j.kittler\}@surrey.ac.uk}
}

\begin{document}
\maketitle
\begin{abstract}
Advanced image fusion methods mostly prioritise high-level missions, where task interaction struggles with semantic gaps, requiring complex bridging mechanisms.
In contrast, we propose to leverage low-level vision tasks from digital photography fusion, allowing for effective feature interaction through pixel-level supervision.
This new paradigm provides strong guidance for unsupervised multimodal fusion without relying on abstract semantics, enhancing task-shared feature learning for broader applicability.
Owning to the hybrid image features and enhanced universal representations, the proposed GIFNet supports diverse fusion tasks, achieving high performance across both seen and unseen scenarios with a single model.
Uniquely, experimental results reveal that our framework also supports single-modality enhancement, offering superior flexibility for practical applications.
Our code will be available at \url{https://github.com/AWCXV/GIFNet}.
\end{abstract}    
\section{Introduction}
\label{sec:intro}

Image fusion combines critical information from multiple sources to produce an output that is more informative and contextually rich, enhancing both human visual interpretation and the performance of downstream computer vision tasks~\cite{liu2018pixelLevelImageFusion, tang2024generative}.
This technique has been shown to be valuable in remote sensing~\cite{javan2021pansharpReview}, medical imaging~\cite{azam2022Medical_IF7_review}, and related fields~\cite{tang2024generative, xu2020accelerated}.
Typically, image fusion is divided into multi-modal fusion and digital photography fusion, based on the properties of the source images~\cite{karim2023current, li2025contifuse}.
Multi-modal fusion combines complementary information from different sensors, such as Infrared and Visible Image Fusion (IVIF), where the infrared channel highlights targets against backgrounds and visible images convey texture details.
Due to the lack of Ground Truth in such tasks, unsupervised approaches are commonly employed~\cite{zhang2023IVIF_TPAMI_review, tang2023exploring}.
In contrast, digital photography fusion, such as Multi-Focus Image Fusion (MFIF) and Multi-Exposure Image Fusion (MEIF), addresses degradations caused by the limitations of the depth of field and inappropriate exposure within a single image~\cite{zhang2021MFIF_TPAMI_review}.
For these tasks, Ground Truth data can be generated artificially, for instance, by synthetically  blurring regions or adjusting exposure, making supervised learning feasible.

\begin{figure}
\centering
\includegraphics[clip,width=\linewidth]{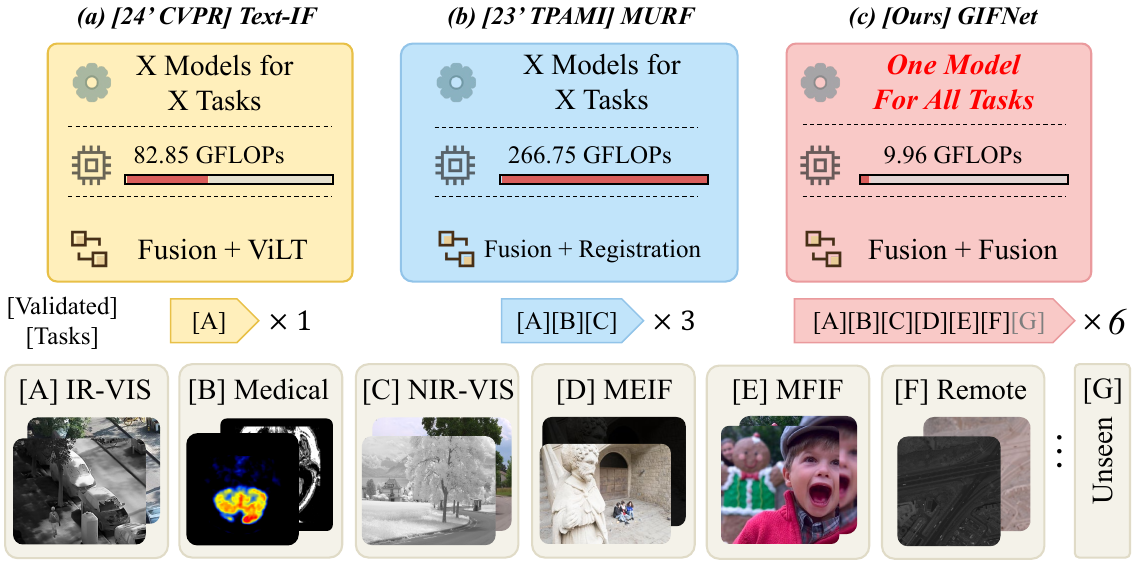}
\vspace{-4mm}
\caption{
A comparison of the versatility and efficiency of advanced multi-task fusion methods.
The indices in the arrows are the fusion tasks validated by the corresponding methods.
}
\label{figure_comparison_gen_effi}
\end{figure}

Currently, a task-interaction mechanism is widely used in advanced image fusion methods.
A typical approach adopted by advanced multi-modal image fusion research is to draw on high-level visual tasks for supervisory signals to direct the fusion.
As a downstream task, these high-level models are usually equipped with a fusion model and receive the fused image as input for iteratively optimising the fusion model and high-level models.
High-level tasks, such as object detection or semantic segmentation~\cite{liu2022target,tang2022SeAFusion,zhang2022watch}, introduce abstract semantic information that can be used to guide task-specific feature learning and improve the fusion performance.
Meanwhile, the improved scene representation derived from the fused image also helps to realise the high-level task better by virtue of the mutual reinforcement mechanism established in this way.

However, such high-level supervision 
is somewhat divorced from the underlying image fusion problem.
As shown in Fig.~\ref{figure_comparison_gen_effi}, when addressing a different image fusion task, this indirect formulation requires training a new fusion model that is tailored to specific features of each task. 
This additional requirement can hinder the deployment of image fusion algorithms on small footprint devices, \textit{e.g.}, mobile phone, which have limited computational resources, as every application requires a different model to solve it.

\begin{figure}[t]
\centering
\includegraphics[clip,width=1\linewidth]{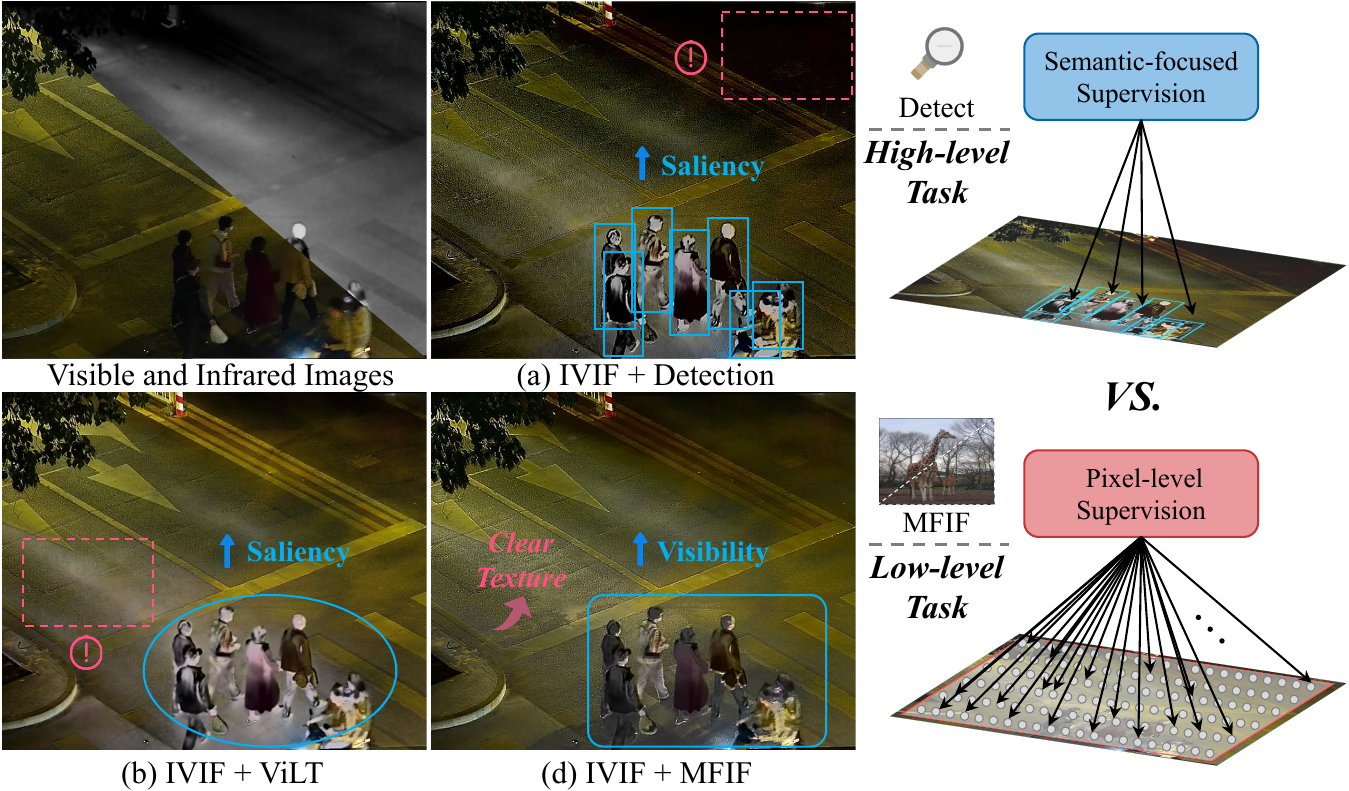}
\vspace{-4mm}
\caption{
Comparison of advanced multi-task fusion methods relying on high-level tasks and the proposed low-level task interaction paradigm.
These semantic-focused paradigms cannot consistently ensure the robust fusion quality as our paradigm does, which provides the pixel-level supervision and presents clear texture details.
}
\label{figure_abstract_comp}
\vspace{-3mm}
\end{figure}

Moreover, the semantic gap between a high level downstream task and the low level image fusion renders the high level supervision signal less than optimal for guiding pixel-focused fusion learning, as it tends to encode features related to object categories, shapes, and scene layouts, rather than fine-grained image details.
This mismatch results in unwanted reliance on complex bridging modules~\cite{xu2023murf} or computationally intensive pre-trained models~\cite{cheng2024textfusionunveilingpowertextual} for different fusion contexts (Fig.~\ref{figure_comparison_gen_effi} (a) and (b)), both of which are resource-heavy and fail to generalise effectively across various fusion contexts.
Also, as shown in Fig.~\ref{figure_abstract_comp}, paradigms (a)~\cite{liu2022target} and (b)~\cite{yi2024textIF} achieve increased saliency for detected objects or text-prompted regions, yet the fused images are often unable to consistently maintain high visual quality, which is significant for other fusion tasks.
We argue that, this phenomenon can be attributed to the absence of pixel-level supervision.

In this work, we aim to address these limitations by promoting cross-task interaction without relying on high-level semantics.
Instead, we use low-level digital photography fusion tasks as a more natural alternative for providing supervisory signals.
Digital photography fusion shares its characteristics with multi-modal fusion, emphasising the preservation of details and focusing on pixel-level feature alignment, rendering it better equipped for enhancing task-shared image features without the semantic mismatch inherent to interacting with high-level tasks.

To this end, we introduce the Generalised Image Fusion Network (GIFNet), a three-branch architecture that supports low-level task interaction for effective fusion.
GIFNet comprises a main task branch, an auxiliary task branch, and a reconciling branch. 
The main and auxiliary task branches, which alternately focus on the multi-modal and digital photography features, promote effective cross-task interaction.
While the reconciling branch, centred on a shared reconstruction task, encourages the network to learn a universal feature representation~\cite{Zhao_2024_CVPR}.
This branch harmonises the optimisation directions of the multi-modal and digital photography branches, preventing divergent task-specific adaptations.
Additionally, our model incorporates a cross-fusion gating mechanism that iteratively refines each task-specific branch, integrating multi-modal and digital photography features to deliver the fusion result.
To minimise the data domain gap between multi-modal and digital photography tasks, we create an RGB-based joint dataset based on the augmentation technique.
With the shared RGB modality derived from identical scenes, the proposed model can focus on consistent feature extraction across the adopted tasks in a unified context, thereby harmonising the training process.

Finally, as shown in Fig.~\ref{figure_comparison_gen_effi} (c), the limited computational cost of low-level fusion tasks in GIFNet reduces GFLOPs by more than 96\% compared to the advanced image fusion method.
Unlike current approaches that focus heavily on high-level vision tasks, prioritising a single category fusion, GIFNet’s integration of both multi-modal and digital photography tasks broadens its applicability across various fusion scenarios with a single model (task-agnostic image fusion).
Besides, rather than amplifying task-specific features, our low-level task interaction enhances task-shared foundational features that are crucial for general image processing, allowing GIFNet to function as a versatile enhancer even for single-modality inputs.
The main contributions of the proposed method include the following:
\begin{itemize}
\item We uniquely demonstrate that collaborative training between low-level fusion tasks, a strategy whose importance was previously not recognised, yields significant performance gains by harnessing cross-task synergies.
\item The reconstruction task and an augmented RGB-focused joint dataset are introduced to align features of different fusion tasks and to address the data support.
\item Our method significantly enhances the versatility of the fusion system, eliminating the need for time-consuming task-specific adaptation.
\item GIFNet pioneers the integration of image fusion and single-modality enhancement processes, extending the scope of image fusion models beyond the multi-modal domain.
\end{itemize}

\section{Related work}
\label{sec:relatedwork}

\subsection{Image Fusion and Downstream Tasks}

With the increased interest in learning-based image fusion, mainstream approaches aim to improve fusion performance by introducing high-level semantics to multi-modal image fusion tasks~\cite{tang2022SeAFusion,zhao2023metafusion,Zhao_2024_ICML}.
This paradigm can enhance the performance of downstream multimodal tasks using the improved fusion results~\cite{liu2022target}.
Additionally, these methods achieve promising objectively measured performance in various image fusion assessments.

However, the reliance on labels from downstream detection or segmentation tasks makes the performance gains costly and compromises their relevance for new fusion tasks.
Besides, the computational burden incurred by involving a relatively large high-level vision model in a low-level image processing technique seems an inappropriate use of resources.
The semantic gap between the features required for image fusion and high-level visual tasks also impairs the quality of fused images~\cite{zhao2023metafusion}.
In FusionBooster~\cite{cheng2024fusionbooster}, Cheng~\textit{et.al.} identify this discrepancy and propose using an enhancer designed specifically to avoid the injection of incompatible semantic information.
Despite the significant performance improvements, this boosting paradigm requires extra training for each fusion task.
Additionally, the commonalities among different fusion tasks are ignored, and the potential of specific features derived from different missions is not fully exploited.

Inspired by this analysis, we propose GIFNet.
It only combines the low-level vision tasks to establish the cross-task interaction and diverse image fusion tasks are used to extract foundational and targeted features.
Thanks to the carefully designed task combination, scenario-specific cooperation mechanisms aimed to reduce the semantic gaps are not required in GIFNet, enabling a more effective multi-task learning paradigm.

\subsection{Generalised Image Fusion Methods}
Some existing studies also aim to develop a generalised image fusion method that performs well across various fusion tasks with different input modalities or image types.
In U2Fusion~\cite{xu2022u2fusion}, Xu~\textit{et.al.} proposed a unified image fusion method based on continual learning, capable of handling multiple fusion tasks.
This work addresses conflicts among different fusion subtasks but fails to promote task interaction during training.
Subsequently, more algorithms have been developed simultaneously to improve image fusion performance and generalisation ability, including CNN-based methods~\cite{zhang2021sdnet,cheng2023mufusion}, Transformer-based solutions~\cite{ma2022swinfusion,zhu2024cvprTaskCustomized}, Mamba-based algorithms~\cite{xie2024fusionmamba}, and some frequency-based approaches~\cite{zhou2024frequency,huang2024efficientFreq}.
However, these paradigms rely heavily on extensive training data spanning diverse fusion tasks and cannot achieve true generalisation without further training on task-specific datasets.
Typically, they depend on multiple image fusion models or specific fusion rule designs~\cite{cheng2021unifusion} to manage different fusion tasks effectively.

In our work, we address this limitation by designing a cross-fusion gating mechanism, involving only the interaction of two representative image fusion tasks from the multi-modal image fusion and the digital photography image fusion.
The learned hybrid image features and enhanced low-level representations enable us to use a single model for achieving task-independent and generalised image fusion.
\section{The Proposed GIFNet}
\label{sec:app}

\subsection{Formulation}
The image fusion paradigm can generally be defined as:
\begin{equation}
    I_{\textrm{f}} = F(I_1,I_2),
\end{equation}
where $I_1$ and $I_2$ are input images, $F$ denotes an image fusion approach, and $I_\textrm{f}$ is the fused image.
Recent methods often incorporate semantic information from high-level vision tasks for the multi-modal image fusion (IVIF) model~\cite{liu2022target,cheng2024textfusionunveilingpowertextual,Zhao_2024_ICML}, aiming to improve performance.
However, this paradigm raises risks of degraded image quality, increased computational cost, and limited generalisation (Fig.~\ref{figure_comparison_gen_effi} and Fig.~\ref{figure_abstract_comp}).

We propose a novel approach by introducing two innovative ideas.
The first is a cross-task interaction mechanism that leverages low-level processing operations across various fusion tasks.
Specifically, we use digital photography image fusion tasks to provide additional task-specific features and supervision signals for the unsupervised IVIF task, thereby improving the generalisation and robustness of the fusion model.
We select Multi-Focus Image Fusion (MFIF) as a representative example of digital photography fusion to demonstrate our GIFNet model, as it performed best among available fusion tasks in our interaction ablation experiments (Sec.~\ref{SecAblation}).

The second innovative feature of our method is the incorporation of single-modality image enhancement capability. 
Introducing digital photography fusion tasks (one image with different settings), the model learns to enhance features without relying on multi-modal input.
By setting both inputs to the same image, we simulate a fusion-like enhancement, focusing on refining details within the single image.
This inference process is formulated as:
\begin{equation}
\hat{X} = F(X,X),
\end{equation}
where $X$ denotes the single modality input, $\hat{X}$ is the enhanced output.
Applications of existing image fusion methods are only restricted to the multi-modal scenarios.
While this new feature enables us to take advantage of the enhanced results for boosting mainstream RGB vision tasks.

\begin{figure*}[t]
\centering
\includegraphics[clip,width=1\linewidth]{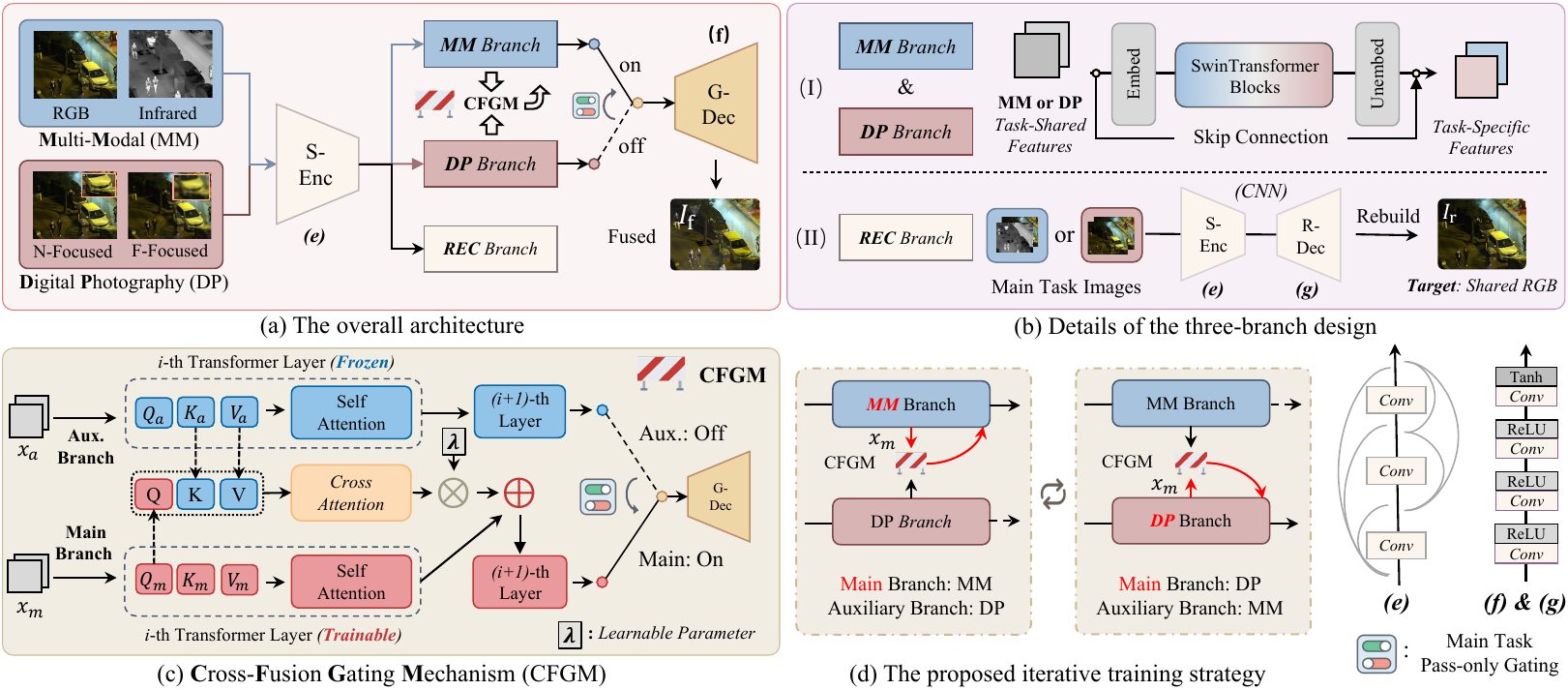}
\vspace{-5mm}
\caption{The network architecture and training process of GIFNet.
As shown in diagram (d), the Multi-Modal (MM) and Digital Photography (DP) branches of our model are trained alternately, based on the specifically designed cross-fusion gating mechanism (c).
}
\label{figure_networkarchitecture}
\vspace{-3mm}
\end{figure*}
\subsection{Measures to Mitigate the Domain Gap and the Task Differences}
\label{secDomainGap}
Our multi-task learning framework requires the model to extract and learn distinct features from input images for each task.
Without taking explicit measures, this diversity can misalign the model’s learning objectives, making it challenging to develop a unified representation that performs effectively across all tasks.

To address this issue, we employ a data augmentation technique to generate an RGB-focused joint dataset from an IVIF benchmark~\cite{jia2021llvip}.
This augmented dataset consists of aligned RGB, infrared, far-focused and near-focused images.
The multi-focus data is obtained by partially blurring clear RGB images (\textit{details are provided in the supplement}).
Since the data is derived from the same scene within a single dataset, the domain gap is effectively reduced.
In addition, we introduce a reconstruction (REC) task in the cross-task interaction.
The REC task facilitates feature alignment across different tasks by focusing on features that are beneficial universally.
This approach ensures that features learned for one task remain relevant and compatible with other tasks, promoting a more coherent and effective interaction among tasks.

\subsection{Model Architecture}
Contemporary image fusion methods often struggle with collaborative learning due to their monolithic network designs, where multiple tasks depend on a singular encoder-decoder structure~\cite{zhang2021sdnet,deng2024mmdrfuse,cheng2023mufusion}.
To address this, our framework introduces a three-branch architecture (as illustrated in Fig.~\ref{figure_networkarchitecture} (a)), which decouples the feature extraction process and facilitates interaction between low-level tasks.
In our model, only the foundational feature extraction part is shared across different tasks.

By focusing on the interaction among low-level tasks, our approach allows task-specific features to be combined directly within the network, removing the need for additional modules to bridge feature or semantic gaps.
This interaction occurs between the Multi-Modal (MM) and Digital Photography (DP) branches, where a cross-task mechanism alternates the roles of main and auxiliary branches (Fig.~\ref{figure_networkarchitecture} (d)).
A gating module then selectively routes the main branch’s hybrid features to the global decoder (G-Dec) for delivering fusion results.
The reconstruction (REC) branch supports this process by extracting task-agnostic features.

Reconstruction Branch:
As shown in Fig.~\ref{figure_networkarchitecture} (b) (II), the REC branch employs an autoencoder to derive universal features from various image fusion tasks. 
By targeting the common RGB modality within our augmented data for the reconstruction, we ensure the effective extraction of task-shared features.
Dense connections in the shared encoder (S-Enc) maximize the feature utilisation, enabling the transmission of the original visual signals to the other branches.

Cross-Fusion Gating Mechanism:
After obtaining these shared features, the MM and DP branches proceed to extract task-specific features of different fusion types(Fig.~\ref{figure_networkarchitecture} (b) (I)).
The proposed Cross-Fusion Gating Mechanism (CFGM) serves as the core technique for controlling these branches, enabling them to fuse task-specific features and stabilise cross-task interaction adaptively. 
In view of its well-known robust global feature extraction ability and its success in capturing task-aware features~\cite{ma2022swinfusion,li2024crossfuse}, we use the efficient SwinTransformer block~\cite{Liu_2021SwinTransformer} to formulate the CFGM.

Within the CFGM, main and auxiliary branches are alternately trained by updating one while freezing the other (Fig.~\ref{figure_networkarchitecture} (c)).
In each training step, we have:
\begin{align}
    \hat{x_m}&  = \textrm{Self\text{-}Att}(x_m), \\
    x_m&  = \hat{x_m} + \lambda\cdot \textrm{Cross\text{-}Att}(\hat{x_m},x_{a}),
\end{align}
where $x_m$ and $x_a$ represent the main and auxiliary task representations, respectively, and $\lambda$ is a learnable parameter that controls the degree of auxiliary task influence. 
$\textrm{Self\text{-}Att}$ and $\textrm{Cross\text{-}Att}$ denote the self attention and cross attention operations.
The interaction is confined to the odd layers to avoid interfering with the SwinTransformer's window shift operation~\cite{Liu_2021SwinTransformer}.


\begin{figure}[t]
\centering
\includegraphics[clip,width=1\linewidth]{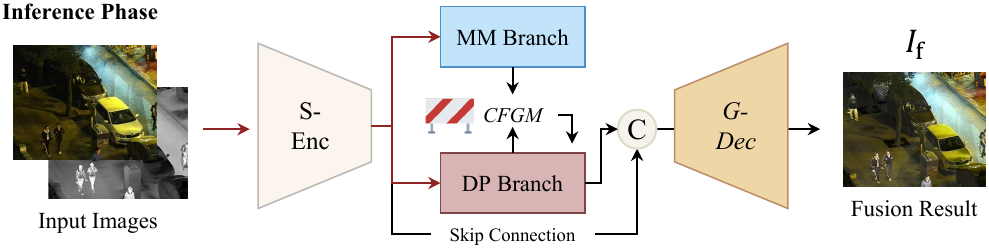}
\caption{An illustration of the inference phase of our GIFNet.
In this stage, only one pair of images will be used to produce the multi-modal and digital photography features.
}
\vspace{-2mm}
\label{figure_inference}
\end{figure}

\subsection{Training and Inference}
In the training process, we adopt two loss functions, \textit{i.e.}, the public loss $\mathcal{L}_{\textrm{pub}}$ and the private loss $\mathcal{L}_{\textrm{pri}}$, defined by the outputs of the REC branch $I_\textrm{r}$ and the fused image $I_{\textrm{f}}$.
The total loss for each task branch is:
\begin{equation}
\label{eqTotalLoss}
    \mathcal{L}_{\textrm{total}} = \mathcal{L}_{\textrm{pub}}+\mathcal{L}_{\textrm{pri}}.
\end{equation}
The public loss $\mathcal{L}_{\textrm{pub}}$ guides the foundational feature extraction by enforcing consistency between the REC branch output and the shared RGB modality ($I_{\textrm{vis}}$):
\begin{equation}
     \mathcal{L}_{\textrm{pub}} = \mathcal{L}_{\textrm{ssim}}(I_r,I_{\textrm{vis}})+\mathcal{L}_{\textrm{mse}}(I_r,I_{\textrm{vis}}),
\end{equation}
where $\mathcal{L}_{\textrm{ssim}}$ and $\mathcal{L}_{\textrm{mse}}$ denote the structural similarity loss and mean squared error loss, respectively.
The structural similarity loss is defined as:
\begin{equation}
    \mathcal{L}_{\textrm{ssim}} (I_{X},I_{Y}) = 1-SSIM(I_X,I_Y).
\end{equation}
Here, $SSIM$ denotes the structural similarity metric~\cite{zhou2004SSIM} between two images.

The private loss is uniquely defined for each task.
During the iterative task-interaction process, only the main task branch and its private loss will be optimised while the other branch is frozen.
Note that, the input for the REC branch is always the main task images.
For the MM branch (IVIT task), which requires the fused image to retain the informative content from the input images~\cite{zhang2023IVIF_TPAMI_review,cheng2023mufusion}, we employ an information-weighted loss function.
Based on the visual explanation studies of convolutional networks~\cite{selvaraju2017grad_cam}, the gradients of a feature map indicate how a specific area contributes to the final network decision-making. 
Using a lightweight DenseNet classification network~\cite{huang2017densely}, we determine the mixing proportions $w_{\textrm{vis}}$ and $w_{\textrm{ir}}$:
\begin{equation}
    GradF(X) = \sum \nabla \phi(\textrm{X}),
\end{equation}
\begin{equation}
    [w_{\textrm{ir}},w_{\textrm{vis}}] = \textrm{softmax}(GradF(I_{\textrm{ir}}),GradF(I_{\textrm{vis}})),
\end{equation}
where $\phi(\textrm{X})$ denotes the extracted image features of modality $X$ via the pre-trained DenseNet121 network.
The private loss for the MM branch is then defined as:
\begin{equation}
    \mathcal{L}_{\textrm{pri}}^{\textrm{MM}} = w_{\textrm{ir}}\cdot \mathcal{L}_{\textrm{mse}}(I_{\textrm{f}},I_{\textrm{ir}})+w_{\textrm{vis}}\cdot \mathcal{L}_{\textrm{mse}}(I_{\textrm{f}},I_{\textrm{vis}}).
\end{equation}

For the DP branch (MFIF task), given that the augmented MFIF data is derived from RGB images ($I_{\textrm{vis}}$), we have ground truth for supervised training (Sec.\ref{secDomainGap}).
Hence, the private loss for this branch is formulated as:
\begin{equation}
    \mathcal{L}_{\textrm{pri}}^{\textrm{DP}} = \mathcal{L}_{\textrm{mse}}(I_{\textrm{f}},I_{\textrm{vis}}).
\end{equation}


As shown in Fig.~\ref{figure_inference}, during inference, different from the training process, only one image pair is required for a single fusion task.
We then extract shared image features, use CFGM to fuse the two sets of specific representations, and finally, the global decoder reconstructs the fused image.
\section{Experimental Results}
\label{sec:exp}

\subsection{Experimental Settings}
\textbf{Training:}
During the training process, only the IVIF dataset (training set of the \textit{LLVIP}~\cite{jia2021llvip}) and the corresponding augmented data for the DP task are used.

\noindent\textbf{Evaluation:}
After training, we directly apply the model to various seen and unseen image fusion tasks,~\textbf{\textit{without any adaption or fine-tuning}}.
The tasks and datasets used include: the \textit{LLVIP} and \textit{TNO}~\cite{2014TNO} datasets for the IVIF task, the \textit{Lytro}~\cite{nejati2015mflytrodataset} and \textit{MFI-WHU}~\cite{zhang2021mffgan} datasets for the MFIF task, the \textit{Harvard} dataset~\cite{cheng2023mufusion} for the medical image fusion task, the \textit{VIS-NIR Scene}~\cite{xu2023murf} dataset for the near-infrared and visible image fusion task, the SCIE dataset~\citep{Cai2018medataset} for the multi-exposure image fusion task, and the \textit{Quickbird} dataset~\cite{zhang2020rethinking} for the remote sensing image fusion task.
We also validate the effectiveness of GIFNet on the classification task using the CIFAR100 dataset~\cite{krizhevsky2009cifar100}.

The evaluation metrics for image fusion include two commonly used correlation-based metrics: Visual Information Fidelity (VIF) and Sum of Correlation Difference (SCD)~\cite{ma2020ddcgan}. Additionally, we include non-reference image quality assessments~\cite{cheng2021unifusion}: Edge Intensity (EI) and Average Gradient (AG) to measure the clarity of the fusion results. For the classification task, we use top-1 and top-5 accuracy.

\begin{figure}[t]
  \centering
\includegraphics[clip,width=0.90\linewidth]{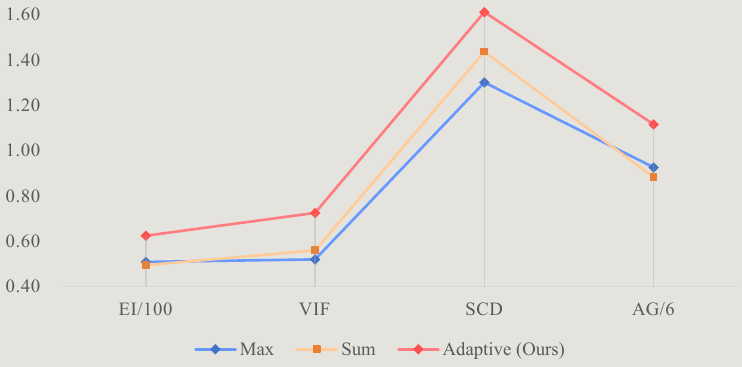}
\vspace{-1mm}
\caption{Quantitative results of the CFGM ablation experiments.
}  
\vspace{-3mm}
  \label{visualisation_cfgm_ablation}
\end{figure}

\begin{figure}[t]
  \centering
\includegraphics[clip,width=1\linewidth]{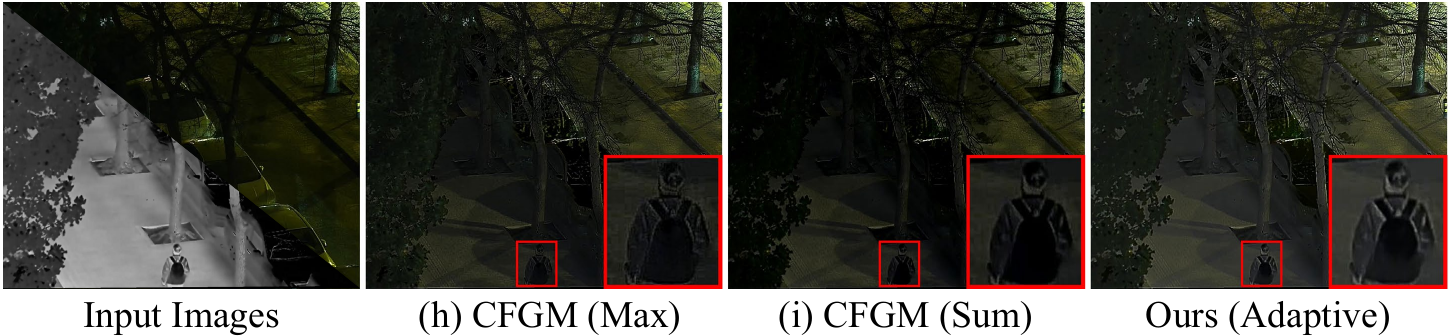}
\vspace{-5mm}
\caption{Visualisation of the CFGM ablation experiments.
}  
\vspace{-2mm}
  \label{Fig_ablation}
\end{figure}

\begin{table}[tbp]
  \centering
  \resizebox{0.95\linewidth}{!}{
    \begin{tabular}{cccccccc}
\cline{1-8}    Case  & MTL   & CFGM  & REC   & EI    & VIF   & SCD   & AG \\
    \hline
    (a)   & IV only & $\times$ & $\times$ & 24.07  & 0.35  & 1.13  & 2.27  \\
    (b)   & MF only & $\times$ & $\times$ & 54.47  & 0.55  & 1.49  & 5.80  \\    
    (c)   & ME only & $\times$ & $\times$ & 24.77  & 0.45  & 0.34  & 2.38  \\
\hdashline    
    (d)   & IV only & $\times$ & \checkmark & 28.71  & 0.53  & 1.45  & 2.71  \\
    (e)   & MF only & $\times$ & \checkmark & 56.14  & 0.59  & 1.50  & 6.00  \\
    (f)   & ME only & $\times$ & \checkmark & 47.17  & 0.61  & 0.69  & 4.97  \\
\hdashline    
    (g)   & IV+MF & $\times$ & $\times$ & -     & -     & -     & - \\
    (h)   &  IV+MF & \checkmark & $\times$ & 49.84  & 0.52  & 1.49  & 5.31  \\
    (i)   & IV+MF  & $\times$ & \checkmark & 52.56  & 0.57  & 1.48  & 5.65  \\
\hdashline       
    (j)   & IV+ ME & \checkmark & \checkmark & 47.69  & 0.56  & 0.42  & 5.08 \\
\cline{3-4}    \rowcolor[rgb]{ .749,  .749,  .749} Ours  & IV+MF & \checkmark & \checkmark & \textbf{62.46 } & \textbf{0.73 } & \textbf{1.61 } & \textbf{6.70 } \\
    \hline
    \end{tabular}%
    }
    \vspace{-1mm}
  \caption{The results of the ablation experiments involving different components and task combinations of the proposed GIFNet (MF and ME refer to the multi-focus and mult-exposure image fusion).}
  \label{table_quantative_ablation}
\vspace{-2mm}
\end{table}%

\subsection{Ablation Experiments}
\label{SecAblation}
In this section, we conduct ablation studies on the IVIF task to demonstrate the efficacy of our GIFNet.
We mainly examined the impact of Multi-Task Learning (MTL) strategy, Cross-Fusion Gating Mechanism (CFGM), and the Reconstruction branch (REC). \textit{More ablation experiments will be provided in the supplement.}

\textbf{Main Components:} As shown in Table~\ref{table_quantative_ablation}, combining single-task training strategy with REC (case (e)), the proposed model already yields impressive results.
However, adding another task without the proposed components would prevent the fusion network from converging, \textit{i.e.}, the fusion network simply produces non-functional outputs (case (g)).
Introducing CFGM or REC independently allows the network to produce valid fusion results (cases (h) and (i)).
The combination of both components optimises cross-task interaction and enhance the feature alignment, leading to the best performance of our GIFNet.

\textbf{Task Combination:} The extra supervision signals from the digital photography task help to enhance the fusion performance.
We further validate this conclusion by employing the supervised MEIF task.
Regarding training data, we use the Information Probe module from the FusionBooster~\cite{cheng2024fusionbooster} to decompose the visible images from the LLVIP dataset, to obtain the overexposed and underexposed images (\textit{examples are provided in the supplementary materials}).
The original visible image is regarded as the GT image.
As depicted in case (j), compared with the single task paradigm (setting(d)), the additional supervised task can consistently improve the performance of multi-modal fusion.
However, the MEIF task, as an auxiliary task, cannot achieve better performance than that of using the MFIF task.
The reason behind this phenomenon may due to the fact that, producing images with higher clarity provides more compatible pixel-level supervision, since there is no conflict between enforcing the fused image to perceive clear content with higher-clarity and preserve as much information as possible.
In contrast, MEIF task only involves the adjusting of the overall exposure degree, which is not always align with the objective of IVIF task.

\textbf{CFGM Module:} Finally, as shown in Fig.\ref{visualisation_cfgm_ablation} and Fig.\ref{Fig_ablation}, replacing the adaptive CFGM strategy (featuring a learnable parameter $\lambda$ for controlling the mixing ratio) with conventional fusion operations demonstrates, both quantitatively and qualitatively, that our adaptive approach provides superior control over the interaction process, yielding more robust fused images.


\begin{figure}[t]
\centering
\includegraphics[clip,width=0.9\linewidth]{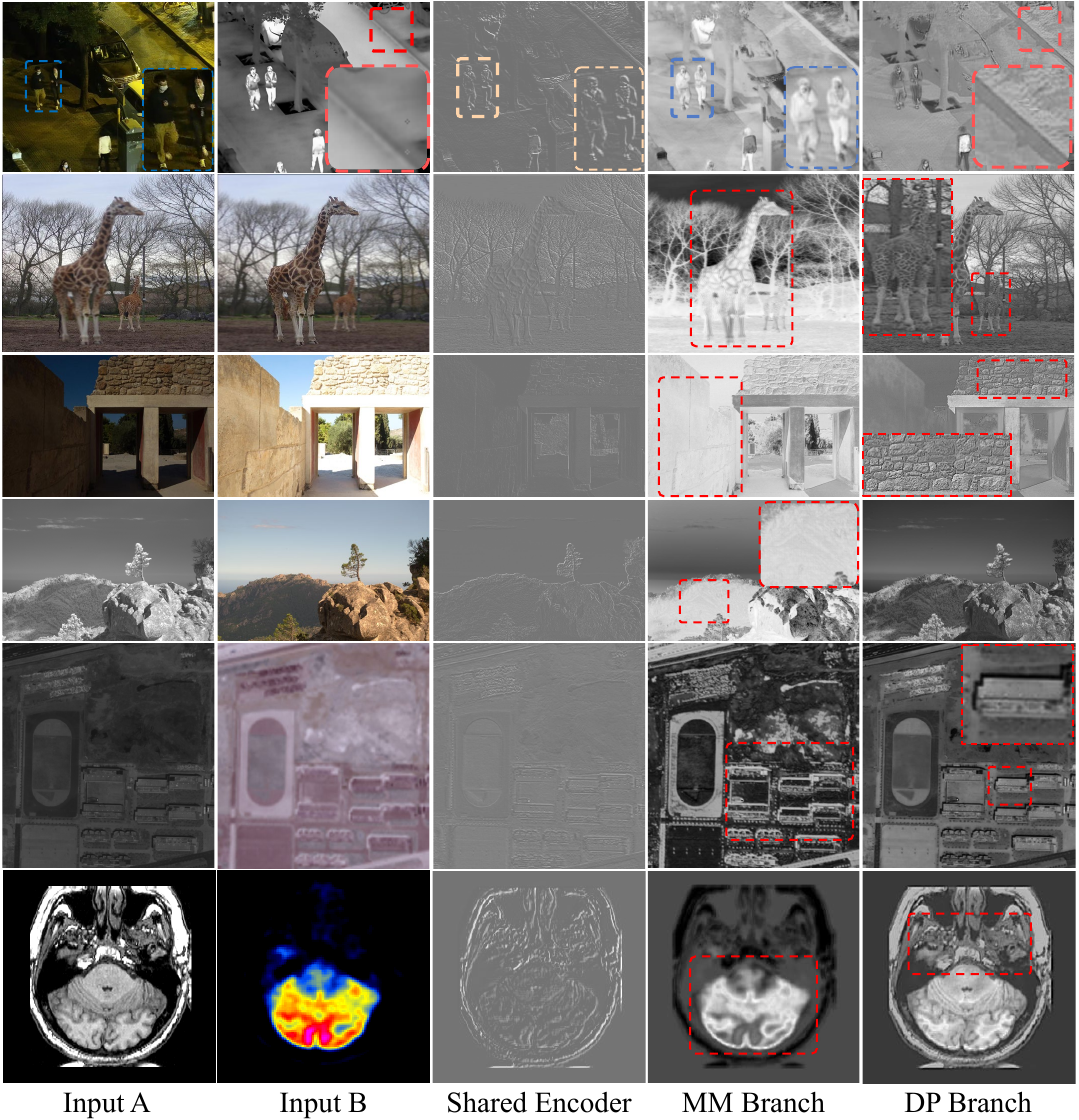}
\vspace{-3mm}
\caption{Visualisations of the feature maps from the shared-encoder and the two branches on various image fusion tasks. 
}
\label{figure_visualisation_featureMaps}
\vspace{-2mm}
\end{figure}

\subsection{Feature Visualisation}
We present visualisations of the feature maps from different components: the shared encoder (S-Enc), the MM branch, and the DP branch, as shown in Fig.~\ref{figure_visualisation_featureMaps}. 
The S-Enc, driven by the image reconstruction objective, captures foundational image features, such as target contours and structural details, which are essential for high-quality image fusion. 

The MM and DP branch visualisations reveal the distinct contributions of each branch to the fusion process.
For instance, in the first case, MM features focus on preserving salient information from the source inputs, such as thermal targets.
Meanwhile, DP features enhance finer details, capturing sharper edges and more defined textures, as well as clearer shadows on the ground.
Similar patterns are observed across other seen and unseen fusion tasks.
Notably, the additional learning of digital photography features consistently benefits various fusion tasks by producing the necessary features for visually robust outputs, as seen in the third example (MEIF task) where enhanced texture details are prominent.

\begin{figure}[t]
\centering
\includegraphics[clip,width=1\linewidth]{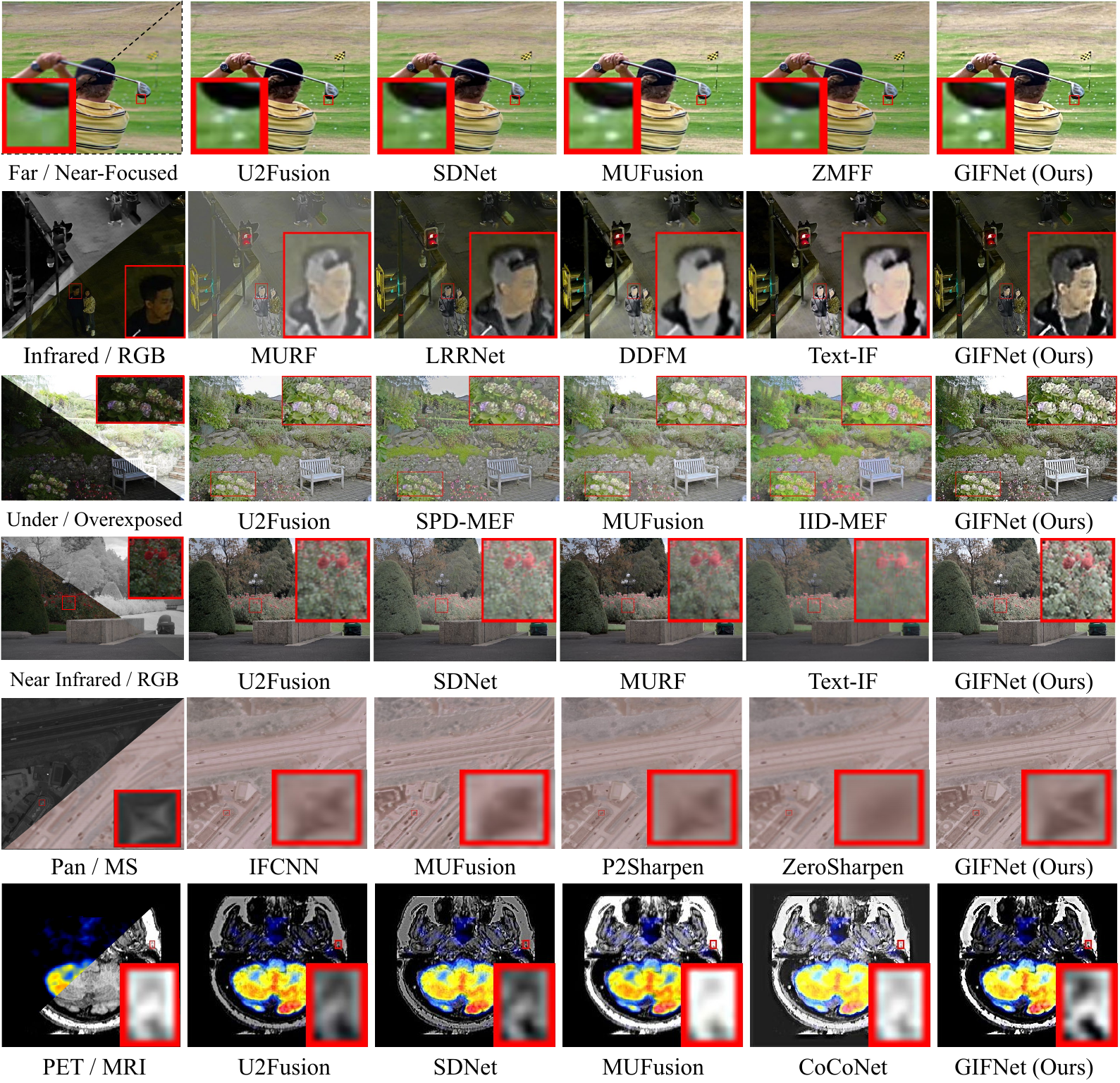}

\caption{Qualitative results of the advanced methods on different fusion tasks.
}
\label{figure_qualitativeALL}
\vspace{-3mm}
\end{figure}

\begin{table*}[tbp]
  \centering

  \vspace{-6mm}
  \resizebox{1\linewidth}{!}{  
    \begin{tabular}{ccccc:cccc|ccccc|ccccc}
    \hline
    \multicolumn{5}{c:}{\textit{(a1) MFIF Task: Lytro}} & \multicolumn{4}{c|}{\textit{(a2) MFIF Task: MFI-WHU}} & \multicolumn{5}{c|}{\textit{(c) MEIF Task: DSCIE}} & \multicolumn{5}{c}{\textit{(d) NIR-VIS Task: Scene}} \\
    \hline
    Method & EI    & VIF   & SCD   & AG    & EI    & VIF   & SCD   & AG    & Method & EI    & VIF   & SCD   & AG    & Method & EI    & VIF   & SCD   & AG \\
    \hline
    U2Fusion & 67.20  & \textbf{1.39 } & 0.84  & 6.34  & \textbf{79.11 } & \textbf{1.50 } & 0.56  & 7.88  & U2Fusion & \textbf{83.00 } & 1.69  & 0.49  & \textbf{8.71 } & IFCNN & 82.13  & 0.90  & 1.14  & 8.72  \\
    UNIFusion* & 70.14  & 1.30  & 0.60  & \textbf{6.77 } & 66.57  & 1.01  & 0.29  & 7.19  & SPD-MEF* & 78.17  & \textbf{1.72 } & 0.49  & 8.12  & U2Fusion & 80.73  & 1.07  & 1.19  & 8.51  \\
    SDNet & 62.98  & 1.12  & 0.75  & 6.16  & 72.98  & 1.16  & 0.63  & \textbf{7.98 } & MEF-GAN* & 80.21  & 1.59  & 0.63  & 8.06  & SDNet & 76.70  & 0.86  & 0.72  & 8.26  \\
    MUFusion & \textbf{70.40 } & 1.34  & \textbf{1.22 } & 6.67  & 77.72  & 1.36  & \textbf{1.11 } & 7.92  & MUFusion & 70.18  & 1.64  & \textbf{0.96 } & 7.19  & MURF*  & 41.71  & 0.41  & 0.19  & 4.22  \\
    ZMFF*  & 58.97  & 1.11  & 0.36  & 5.48  & 57.90  & 1.03  & 0.33  & 5.49  & IID-MEF* & 59.12  & 1.12  & 0.36  & 6.13  & Text-IF & \textbf{88.19 } & \textbf{1.49 } & \textbf{1.45 } & \textbf{9.16 } \\
    \rowcolor[rgb]{ .749,  .749,  .749} GIF (Ours)  & \underline{\textbf{80.86 }} & \underline{\textbf{1.74 }} & \underline{\textbf{1.37 }} & \underline{\textbf{7.71 }} & \underline{\textbf{91.61 }} & \underline{\textbf{1.94 }} & \underline{\textbf{1.29 }} & \underline{\textbf{9.25 }} & GIF (Ours)  & \underline{\textbf{111.27 }} & \underline{\textbf{2.52 }} & \underline{\textbf{1.04 }} & \underline{\textbf{12.00 }} & GIF (Ours)  & \underline{\textbf{101.32 }} & \underline{\textbf{1.51 }} & \underline{\textbf{1.45 }} & \underline{\textbf{10.83 }} \\
    \hline
    \multicolumn{5}{c:}{\textit{(b1) IVIF Task: LLVIP}} & \multicolumn{4}{c|}{\textit{(b2) IVIT Task: TNO}} & \multicolumn{5}{c|}{\textit{(e) Remote Task: QuickBird}} & \multicolumn{5}{c}{\textit{(f) Medical Task: Harvard}} \\
\hline
    Method & EI    & VIF   & SCD   & AG    & EI    & VIF   & SCD   & AG    & Method & EI    & VIF   & SCD   & AG    & Method & EI    & VIF   & SCD   & AG \\    
    \hline    
    MURF  & 38.02  & 0.25  & 0.56  & 3.75  & 45.93  & 0.94  & 1.52  & 4.35  & IFCNN & 18.30  & 1.16  & 0.86  & \textbf{1.73 } & U2Fusion & 73.29  & 0.78  & 1.46  & 7.06  \\
    LRRNet* & 34.93  & 0.34  & 0.95  & 3.64  & 36.37  & 0.75  & 1.40  & 3.75  & TextFusion & 13.73  & 0.76  & 0.32  & 1.29  & IFCNN & \textbf{98.67}  & 0.92  & 1.33  & \underline{\textbf{9.57}}  \\
    DDFM*  & 41.13  & 0.51  & 1.55  & 4.43  & 31.01  & 0.67  & 1.60  & 3.03  & MUFusion & \textbf{18.45 } & 1.05  & -0.02  & 1.72  & SDNet & 83.86  & 0.71  & \textbf{1.60 } & 8.39  \\
    CDDFuse* & 52.32  & \textbf{0.79 } & \textbf{1.58 } & 5.42  & 43.83  & 0.91  & \textbf{1.66 } & 4.55  & P2Sharpen* & 16.66  & \textbf{1.20 } & \textbf{0.94 } & 1.56  & MUFusion & 88.66  & \textbf{0.97 } & 1.23  & 8.39  \\
    Text-IF* & \textbf{61.30 } & \underline{\textbf{0.93 }} & 1.49  & \textbf{6.33 } & \textbf{46.09 } & \underline{\textbf{1.04 }} & 1.53  & \textbf{4.61 } & ZeroSharpen* & 13.20  & 0.94  & 0.48  & 1.26  & CoCoNet*  & 89.55  & 0.71  & 1.04  & 8.84  \\
    \rowcolor[rgb]{ .749,  .749,  .749} GIF (Ours)  & \underline{\textbf{62.46 }} & 0.73  & \underline{\textbf{1.61 }} & \underline{\textbf{6.70 }} & \underline{\textbf{52.30 }} & \textbf{0.99 } & \underline{\textbf{1.66 }} & \underline{\textbf{5.24 }} & GIF (Ours)  & \underline{\textbf{23.02 }} & \underline{\textbf{1.56 }} & \underline{\textbf{1.04 }} & \underline{\textbf{2.16 }} & GIF (Ours)  & \underline{\textbf{100.71 }} & \underline{\textbf{1.10 }} & \underline{\textbf{1.68 }} & \underline{\textbf{9.73 }} \\
    \hline
    \end{tabular}%
}

  \caption{Quantitative results of the dedicated (*) or unified methods on various image fusion tasks. (\underline{\textbf{Bold}}: best, \textbf{Bold}: second best) }
      \label{table_quantative_all}  

\end{table*}%

\subsection{Multiple Modalities - Seen Tasks}
In this section, we present the fusion results of our GIFNet on the tasks related to our training data, \textit{i.e.}, MFIF and IVIF tasks.
We compare the proposed method with dedicated algorithms for these two tasks, including Text-IF~\cite{yi2024textIF}, CDDFuse~\cite{zhao2023cddfuse}, DDFM~\cite{Zhao_2023_ICCV_DDFM}, LRRNet~\cite{li2023lrrnet}, ZMFF~\cite{hu2023zmff} and UNIFusion~\cite{cheng2021unifusion}.
We also compare with generalised image fusion methods, including MURF~\cite{xu2023murf}, MUFusion~\cite{cheng2023mufusion}, U2Fusion~\cite{xu2022u2fusion}, and SDNet~\cite{zhang2021sdnet}.

\textit{MFIF task:}
As shown in Table~\ref{table_quantative_all} (a1) and (a2), our GIFNet achieves promising results in terms of various image fusion assessment metrics.
For example, the best performance in VIF, with an increase of 25\%, demonstrates that our fusion results can effectively enhance the source information, as seen  in the first row of Fig.~\ref{figure_qualitativeALL}.

\textit{IVIF task:}
For the IVIF task, as shown in the second row of Fig.~\ref{figure_qualitativeALL}, our fusion results, benefiting from collaborative training, can better adjust the mixing proportion of source modalities.
The abundant texture details from the RGB image are well-preserved, and the thermal radiation information contributes to a brighter scene appearance. 
As a result, in both low-light and common conditions, GIFNet generally achieves the best performance across all these quantitative experiments (Table~\ref{table_quantative_all} (b1) and (b2)).
The relatively poor results on VIF of the LLVIP dataset can be attributed to the ``choose-max" fusion strategy in CDDFuse and Text-IF, which retains the source content with the higher pixel value from the input. 
While this approach ensures high visual fidelity (VIF), the fused images tend to bias towards one input modality, ignoring the information from the other (see the visualisation of Text-IF)~\cite{cheng2024textfusionunveilingpowertextual}.

\subsection{Multiple Modalities - Unseen Tasks}
In this section, we present the fusion results of our GIFNet on tasks not involved in training, including multi-exposure image fusion, near-infrared and visible image fusion, remote sensing image fusion, and medical image fusion tasks.
Similarly, we further compare our method with the algorithms designed specifically for these four tasks, including MEF-GAN~\cite{xu2020mef}, SPD-MEF~\cite{li2020fastMEFDecomposition},  IID-MEF~\cite{zhang2023iid}, MURF~\cite{xu2023murf}, P2Sharpen~\cite{zhang2023p2sharpen}, ZeroSharpen~\cite{wang2024zero}, CoCoNet~\cite{liu2024coconet},
TextFusion~\cite{cheng2024textfusionunveilingpowertextual}, which incorporates textual information in the image fusion field, and a generalised method IFCNN~\cite{zhang2020ifcnn}.

\textit{MEIF Task:}
Our GIFNet performs well with poorly exposed images in the MEIF tasks. As shown in the third row of Fig.~\ref{figure_qualitativeALL}, in terms of the overall exposure, which is a significant criterion in this task, our result has more appropriate brightness without serious color distortion (see the highlighted regions).
For the quantitative assessment (Table~\ref{table_quantative_all} (c)), compared with advanced approaches, we achieve much higher performance on all image fusion metrics, \textit{e.g.}, VIF (+46.7\%) and AG (+37.8\%).

\textit{NIR-VIS Task:}
This task is similar to IVIF but replaces the mid-far infrared modality with a near-infrared image. As shown in the fourth row of Fig.~\ref{figure_qualitativeALL}, the existing fusion methods consistently improve low-light conditions of RGB images using the information conveyed by the NIR modality, while our GIFNet exhibits the clearest texture details.
The quantitative results also demonstrate that GIFNet outperforms existing algorithms (Table~\ref{table_quantative_all} (d)).
Notably, although MURF is trained on this task, it focuses more on addressing the registration issue, resulting in relatively poor performance.

\textit{Remote Task:}
This task, also known as Pansharpening, aims to simultaneously keep the spatial and spectral resolution of panchromatic and multispectral images. 
As illustrated in the second last row of Fig.~\ref{figure_qualitativeALL}, like previous tasks, GIFNet obtains fused images with sharper edge information and superior imaging quality.
In contrast, competitors fail to maintain the shape of objects from the high-resolution panchromatic modality.
Although specifically designed for this task, P2Sharpen and ZeroSharpen are surpassed by our approach across multiple metrics, as shown by the quantitative results in Table~\ref{table_quantative_all} (e).

\textit{Medical Task:}
The medical image fusion task aims to preserve salient organ structures from Magnetic Resonance Imaging (MRI) and clear functional information from Positron Emission Tomography (PET). 
As shown in Table~\ref{table_quantative_all} (f), despite not being trained specifically for this task, GIFNet demonstrates strong visual information fidelity (VIF) and maintains a high correlation with the source inputs (SCD) in its fusion results.
This performance is consistent with the visualisation in the last row of Fig.~\ref{figure_qualitativeALL}, \textit{i.e.}, with enhanced details, which shows clearly that  the results of GIFNet well present the local structure from the MRI modality.

\begin{table}[tbp]
  \centering

  \resizebox{1\linewidth}{!}{   
    \begin{tabular}{ccccc}
    \hline
    Method & Venue & Task Combination & Top-1 Acc & Top-5 Acc. \\
    \hline
    TarDAL++ & 22' CVPR & IVIF+Detect & 46.62\% & 76.11\% \\
    MURF  & 23' TPAMI & IVIF+Register & 50.04\% & 79.91\% \\
    SDNet & 21‘ IJCV & IVIF  & 50.28\% & 79.62\% \\
    MUFusion & 23' Inf. Fus. & IVIF  & 50.39\% & 79.59\% \\
    SDNet$^{\dagger}$ & 21' IJCV & MFIF  & 50.83\% & 79.56\% \\
    TextFusion & 24' Inf. Fus. & IVIF+ViLT & 51.15\% & 80.81\% \\
    MUFusion$^{\dagger}$ & 23' Inf. Fus. & MFIF  & 51.50\% & 79.86\% \\
    U2Fusion & 22' TPAMI & IV+ME+MF & 51.58\% & 80.38\% \\
    CDDFuse & 23' CVPR & IVIF  & 52.20\% & 80.00\% \\
    Text-IF & 24' CVPR & IVIF+ViLT & 52.57\% & 80.98\% \\
    Cifar-original & -     & -     & 54.11\% & 83.03\% \\
    \rowcolor[rgb]{ .749,  .749,  .749} GIFNet & Ours  & IVIF+MFIF & \underline{\textbf{56.18\%}} & \underline{\textbf{84.95\%}} \\
    \hline
    \end{tabular}%
}

  \caption{The classification results of the ResNet56 when using different data for training. The original CIFAR100 dataset and enhanced data using different image fusion approaches are regarded as the training set. ($\dagger$: this method is trained with a different task)} 

    \label{table_quantative_cifar}
\end{table}%

\begin{figure}[t]
\centering
\includegraphics[clip,width=1\linewidth]{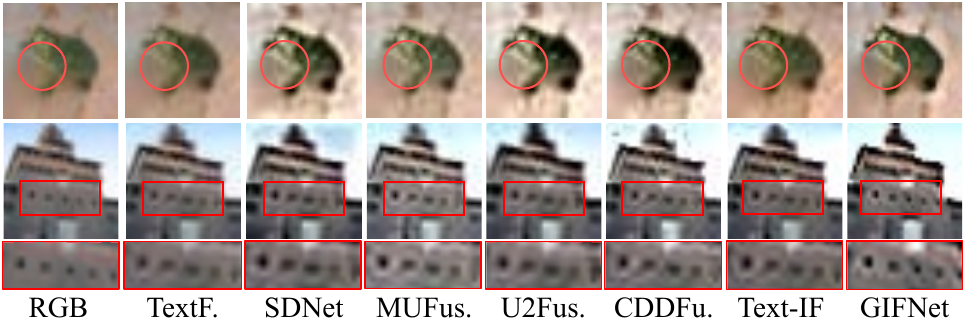}
\vspace{-5mm}
\caption{Visualisations of the advanced image fusion methods on the CIFAR100 single modality enhancement task.
}
\label{figure_qualitativeCifar}
\vspace{-4mm}
\end{figure}

\subsection{Single Modality: the Classification Task}
Our GIFNet’s versatility encompasses both multi-modal image processing and single modality tasks.
This experiment evaluates GIFNet’s ability to boost RGB image classification using enhanced images as inputs~\cite{chen2021hybrid}. 
The original CIFAR100 training set and enhanced data obtained through different image fusion approaches are used to train the ResNet56 network~\cite{he2016resnet} from scratch.
Once trained, the ResNet56 classifier is tasked with evaluating the performance on the original test set.

As illustrated in Fig.~\ref{figure_qualitativeCifar}, we present original CIFAR100 RGB images alongside enhanced versions produced by different approaches.
GIFNet demonstrates a notable improvement in image quality.
For instance, in the first row, the blurring present in the original data is mitigated, with clearer information being preserved.
In the second example, our method excels in edge enhancement, outperforming other techniques.

Quantitative assessments, as shown in Table~\ref{table_quantative_cifar}, indicate that certain fusion methods yield comparable classification performance to the original dataset without improving image quality, such as SDNet and MUFusion. 
Note that, U2Fusion, leveraging even more fusion tasks, suffers from a lack of effective interaction in its sequential training strategy, leading to suboptimal enhancement.
In contrast, using the task-independent representation from the cross-task interaction, GIFNet is the only method to surpass the original training setting.

\section{Conclusion}
\label{sec:con}
This paper introduces a novel approach to low-level task interaction for generalised image fusion, addressing a largely overlooked aspect of the field. By integrating a shared reconstruction task and an RGB-based joint dataset, we effectively reduce task and domain discrepancies, establishing a collaborative training framework. Our model, supported by a cross-fusion gating mechanism, demonstrates superior generalisation and robust fusion performance. Additionally, GIFNet pioneers the application of fusion techniques to single-modality enhancement, representing a significant advancement in the image fusion research.

\textbf{Acknowledgement}: This work is supported by the National Natural Science Foundation of China (62020106012, U1836218, 62106089, 62202205), the 111 Project of Ministry of Education of China (B12018), the Engineering and Physical Sciences Research Council (EPSRC) (EP/V002856/1).

{
    \small
    \bibliographystyle{ieeenat_fullname}
    \bibliography{main}
}


\end{document}